%% file: paper.tex
\documentclass[]{fairmeta}

\input{preamble}
\usepackage{array}
\renewcommand{\arraystretch}{1.3}
\usepackage{hyperref}
\usepackage{url}

\title{\modelname{}: An Auto-regressive Foundation Model for \\ Image Understanding and Generation}

\author[2*\dagger]{Zhiyang Xu}
\author[3*\dagger]{Jiuhai Chen}
\author[1]{Zhaojiang Lin}
\author[4\dagger]{Xichen Pan}
\author[5]{Lifu Huang}
\author[3]{Tianyi Zhou}
\author[1]{Madian Khabsa}
\author[1]{Qifan Wang}
\author[1]{Di Jin}
\author[1\ddagger]{Michihiro Yasunaga}
\author[1\ddagger]{LILI YU}
\author[1]{Xi Victoria Lin}
\author[1]{Shaoliang Nie}

\affiliation[1]{Meta}
\affiliation[2]{Virginia Tech}
\affiliation[3]{University of Maryland}
\affiliation[4]{New York University}
\affiliation[5]{UC Davis}

\contribution[*]{Equal Contribution}
\contribution[\dagger]{Work done during internship at Meta}
\contribution[\ddagger]{Work done while at Meta}

\abstract{\input{sec/0_abstract}}

\date{\today}
\correspondence{Zhiyang Xu at \email{zhiyangxu1@meta.com}, Shaoliang Nie at \email{snie@meta.com}}

\begin{document}

\maketitle

\begin{center}
    \captionsetup{type=figure}
    \includegraphics[width=\textwidth]{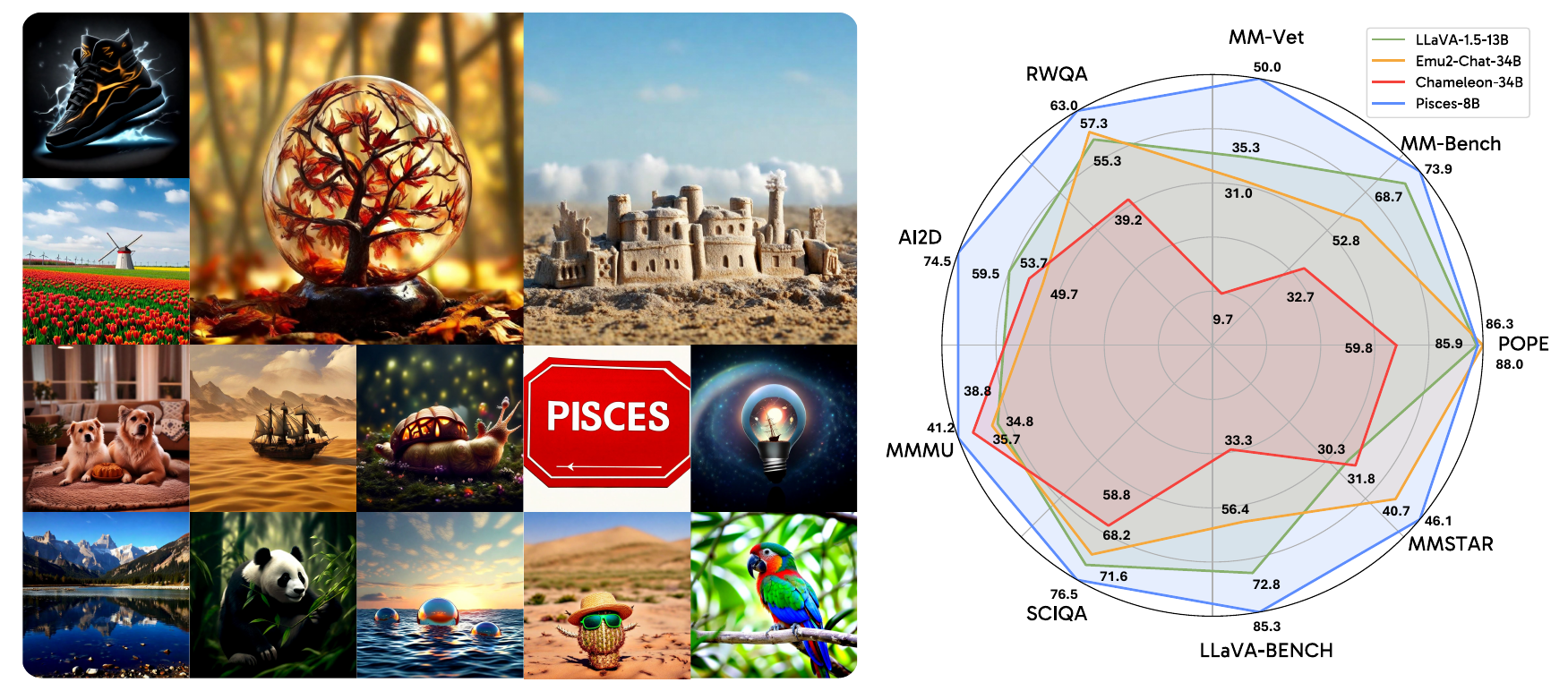}
    \caption{\modelname{} achieves strong performance on both image generation and understanding.}
    \label{fig:teaser}
\end{center}

\input{sec/1_intro}
\input{sec/2_method}

\input{sec/3_experiment}

\input{sec/4_results}

\input{sec/5_related_work}
\input{sec/6_conclusion}

\clearpage
\newpage
\bibliographystyle{assets/plainnat}
\bibliography{paper}



\end{document}

%% file: preamble.tex
\usepackage{graphicx}               
\usepackage{tabularx}               
\usepackage{booktabs}
\usepackage{amsmath}
\usepackage{stmaryrd}

\usepackage{soul}

\newcolumntype{C}{>{\centering\arraybackslash}X}
\usepackage{multirow}               
\usepackage{diagbox}                
\usepackage{hhline}                 
\usepackage{color,colortbl}         
\usepackage{amsmath}                
\usepackage{amssymb}                
\usepackage{mathtools}              
\usepackage{enumitem}               

\usepackage{subcaption}
\usepackage{caption}
\usepackage{booktabs}
\usepackage{extarrows}
\usepackage{wrapfig}
\usepackage{makecell}
\usepackage{fdsymbol}
\usepackage{placeins}

\usepackage{amssymb}
\usepackage{pifont}

\definecolor{RoseQuartzBg}{HTML}{F7CAC9}
\definecolor{ForestGreen}{HTML}{228b22}
\definecolor{RoseQuartz}{HTML}{F5A798}
\definecolor{Serenity}{HTML}{92A8D1}
\definecolor{OrangeRed}{rgb}{1.0, 0.27, 0.0}
\definecolor{Red}{rgb}{1.0, 0.0, 0.0}
\definecolor{forestgreen}{rgb}{0.13, 0.55, 0.13}
\definecolor{Turquoise}{HTML}{0F4C81}
\definecolor{columbiablue}{rgb}{0.61, 0.87, 1.0}
\definecolor{lightblue}{rgb}{0.84, 1.0, 1.0}
\definecolor{lightgreen}{rgb}{0.9, 1., 0.85}
\definecolor{lightpurple}{HTML}{E6E6FA}
\definecolor{Gray}{gray}{0.9}

\usepackage{xparse}
\usepackage{footmisc}
\NewDocumentCommand{\ying}{ mO{} }{\textcolor{teal}{\textsuperscript{\textit{chao}}\textsf{\textbf{\small[#1]}}}}

\NewDocumentCommand{\lifu}{ mO{} }{\textcolor{blue}{\textsuperscript{\textit{Lifu}}\textsf{\textbf{\small[#1]}}}}

\NewDocumentCommand{\zhiyang}{ mO{} }{\textcolor{Turquoise}{\textsuperscript{\textit{zhiyang}}\textsf{\textbf{\small[#1]}}}}

\newcommand*\modelname{\textsc{Pisces}}

\newcommand*\arch{{\text{decoupled visual encoding architecture}}}

\definecolor{deepblue}{rgb}{0.0, 0.0, 0.55}



%% file: sec/0_abstract.tex
\begin{abstract}

Recent advances in large language models (LLMs) have enabled multimodal foundation models to tackle both image understanding and generation within a unified framework. Despite these gains, unified models often underperform compared to specialized models in either task. 
A key challenge in developing unified models lies in the inherent differences between the visual features needed for image understanding versus generation, as well as the distinct training processes required for each modality.
In this work, we introduce \modelname{}, an auto-regressive multimodal foundation model that addresses this challenge through a novel \arch{} and tailored training techniques optimized for multimodal generation. Combined with meticulous data curation, pretraining, and finetuning, \modelname{} achieves competitive performance in both image understanding and image generation.
We evaluate \modelname{} on over 20 public benchmarks for image understanding, where it demonstrates strong performance across a wide range of tasks. Additionally, on GenEval, a widely adopted benchmark for image generation, \modelname{} exhibits robust generative capabilities.
Our extensive analysis reveals the synergistic relationship between image understanding and generation, and the benefits of using separate visual encoders, advancing the field of unified multimodal models.
\end{abstract}

%% file: sec/1_intro.tex
\section{Introduction}
While multimodal foundation models focusing on either image understanding or image generation have been widely studied and exhibit strong performance, unified models capable of excelling in both tasks remain underexplored. Recent works~\cite{cm3, gill,nextGPT, dreamllm, anyGPT, Emu2,ge2024seed, chameleonteam2024chameleon, wang2024emu3nexttokenpredictionneed, moss} have made initial efforts towards this goal but a significant performance gap persists between unified multimodal models and specialized multimodal models, limiting the practicality of these models in real-world applications.

Prior studies \cite{liu2023llava, li2023blip, zhu2023minigpt, Dai2023InstructBLIPTG, liu2023improved}, have shown that training an image understanding model by leveraging pretrained large language models (LLMs) \cite{T5, llama,llama2} and vision encoders \cite{clip} delivers strong performance while significantly reducing computational costs compared to training models from scratch \cite{alayrac2022flamingo, wang2022ofa, BEIT3}. Following their path, we opt to design the architecture of \modelname{} based on pretrained large-language models \cite{llama3.1}, CLIP image encoders \cite{clip, evaClip} and diffusion models \cite{latentDiffusion, sdxl, SD3}, optimizing the use of pretrained components.
Specifically, we utilize a pretrained CLIP encoder to transform images into continuous visual representations, which serve dual roles in our framework i.e., providing context for image understanding and acting as supervision for image generation.
The diffusion model behaves as an image decoder and is trained to decode the images vectors encoded by the CLIP model into the original images.

However, both prior research and our pilot studies in Section~\ref{abl:decouple_encoder} and \ref{abl:num_token} highlight significant limitations with this approach \cite{ge2024seed}. Image understanding benefits from higher input resolutions, which necessitate a longer sequence of image vectors (e.g., 768 or 4096 vectors) to capture the full detail of an image \cite{liu2023improved, llava_UHD}. Conversely, for image generation, autoregressively generating such a long sequence of consistent visual vectors proves challenging for LLMs. In practice, a shorter sequence of 32 or 64 vectors is sufficient to recover most visual details of the original image and is easier for the model to learn \cite{Emu2, ge2024seed, chen2024deepcompressionautoencoderefficient}. Additionally, it is usually the case that the best publicly available image encoders for understanding~\cite{siglip} and generation~\cite{evaClip} are not the same model.
To address this challenge, we propose a novel \arch{} for image understanding and generation, as shown in Figure~\ref{fig:arch}. This architecture allows each task to utilize distinct image encoders, projection layers, and tailored visual vector lengths, enhancing model design flexibility while reducing inference costs relative to a single-encoder setup. Consequently, for image understanding \modelname{} can reason over a long sequence of visual vectors with rich visual details while enjoying the better visual quality and high efficiency of modeling a short sequence of image vectors for generation.

We introduce a three-stage training process that progressively enables novel capabilities in multimodal generation for \modelname{}. In the first stage, we pretrain \modelname{} on high-quality image and short caption pairs sourced from Shutterstock, allowing the model to learn foundational skills in image and caption generation. 
In the second stage, we continue pretraining the model on image and detailed-caption pairs, enabling fine-grained alignment between visual features and textual tokens in the multimodal generation.
In the third stage, we further finetune the model on a meticulously curated instruction-tuning dataset, where each instance includes a user instruction paired with either a textual response or an image output. This dataset spans a wide range of downstream tasks and incorporates diverse user instructions, ensuring the model's strong instruction-following capability and robustness across varied scenarios.

In our comprehensive evaluation, we first assess \modelname{}'s performance on over 20 public benchmarks for image understanding, demonstrating that \modelname{} achieves superior results on most of them, even surpassing models specifically designed for image understanding. Second, we report \modelname{}'s performance on GenEval, a popular image generation benchmark, where it demonstrates strong image generation capability and instruction following capability.
Additionally, we uncover a synergistic relationship between image generation and understanding tasks. Surprisingly, training these two tasks together within a unified multimodal framework shows that image understanding tasks can significantly enhance image generation performance, and conversely, image generation can benefit image understanding performance. We also conduct ablation study to highlight the benefits of using separate vision encoders for image understanding and generation. These insights advance the future research in the field of unified multimodal models.

%% file: sec/2_method.tex
\section{Model Architecture}

\begin{figure}[t]
  \centering
   \includegraphics[width=0.85\linewidth]{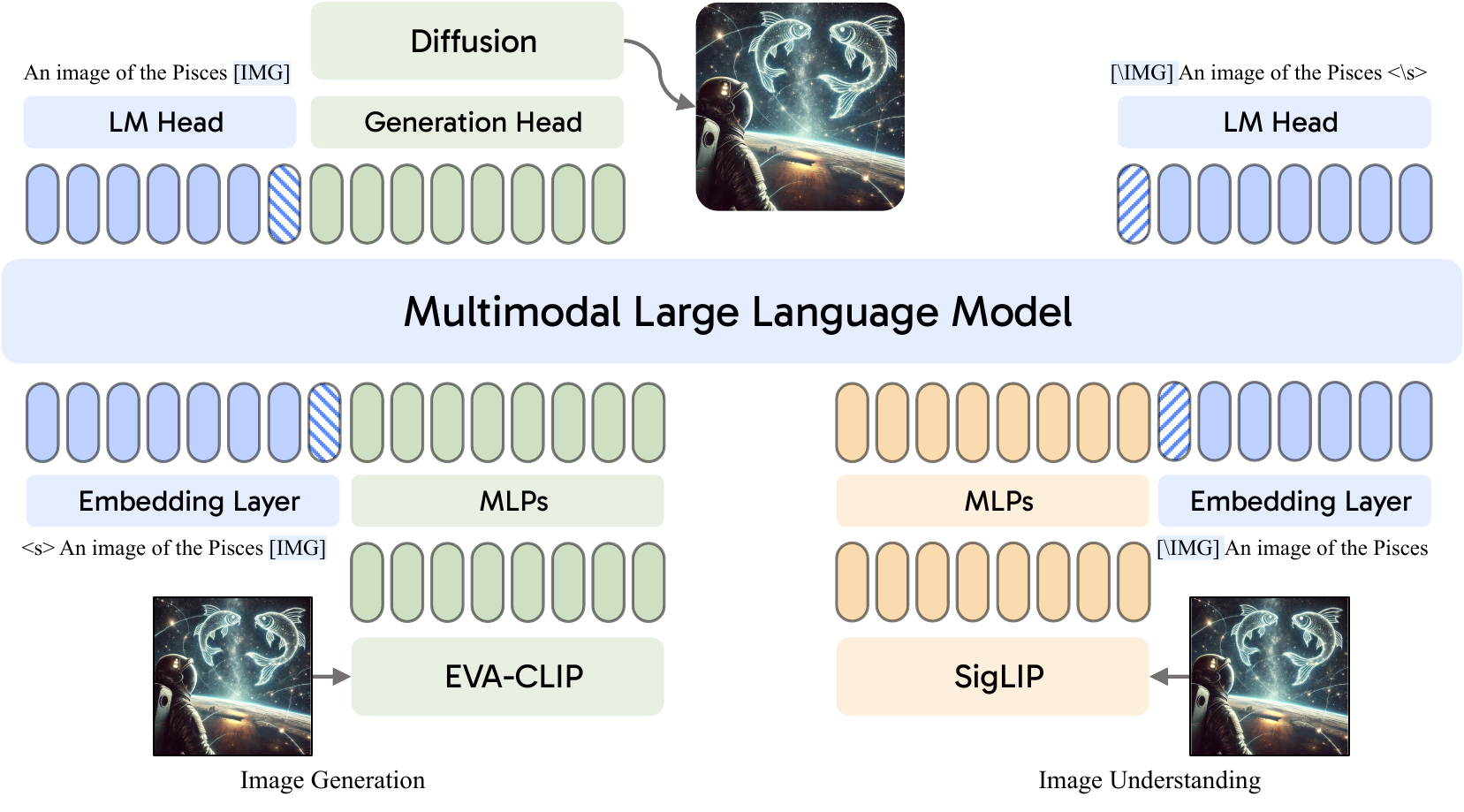}
   \caption{The main architecture of \modelname{}. On the left, we present the architecture for image generation and the right side depicts the architecture for image understanding. Two tasks utilize different visual encoders and projection layers but share the same multimodal language model.}
   \label{fig:arch}
   \vspace{-5mm}
\end{figure}

Our model design is inspired by the architectures of recent multimodal models~\cite{liu2023llava,Emu2,ge2024seed}, which effectively utilize pretrained models to achieve strong performance. As shown in Figure \ref{fig:arch}, we present a novel \arch{} for an autoregressive multimodal model capable of both image understanding and generation. The model includes a pretrained LLM, an image encoder \(\phi\) dedicated to image understanding, a separate image encoder \(\varphi\) optimized for image generation, and a diffusion model. We describe each component of our model in the following sections.

\subsection{Decouple Visual Representations}
The design philosophy of the decoupled visual encoding architecture is rooted in the intrinsic differences between the visual representations needed for image understanding and image generation. Specifically, image understanding demands that the visual encoder extracts detailed and rich semantic information from raw images to facilitate accurate analysis and comprehension, necessitating a long sequence of image vectors. In contrast, image generation requires the visual encoder to compress pixel-level information into a compact sequence of vectors, efficiently capturing the essence of the visual appearance while being optimized for autoregressive generation. Below we explain the details of the separate visual encoding process.

\paragraph{\textbf{Image Understanding}} 
Given an input image $\mathcal{I}$, the image-understanding vision encoder $\phi$ processes it into a sequence of continuous image representation $\phi(\mathcal{I})$. These representations are then projected into the language model's latent space through an MLP module, resulting in 
\[\mathcal{V}_n=\text{MLPs}(\phi(\mathcal{I})) \in \mathbb{R}^{n \times d}
\], where $n$ denotes the number of visual tokens, $d$ is the hidden dimension of the visual tokens after MLP projection. 

\paragraph{\textbf{Image Generation}}
The image-generation visual encoder \(\varphi\) processes the image \(\mathcal{I}\) into a sequence of continuous vectors, represented by \(\varphi(\mathcal{I})\). In the image generation task, the LLM is trained to predict these vectors in an autoregressive manner. However, autoregressively generating such an extended sequence of continuous image vectors poses significant challenges for LLMs (see section \ref{abl:num_token} for further analysis). 
To mitigate this, we utilize average pooling to decrease the number of visual tokens, thereby making the sequence more manageable while preserving crucial visual information.
To achieve this, the sequence of image tokens \(\varphi(\mathcal{I})\) is first reshaped into a 2D structure. Next, we apply 2D pooling with a stride of \(a\) to downsample this structure, resulting in a flattened sequence of \(m\) tokens after pooling. Finally, additional MLP layers project these pooled image vectors into the latent space of the LLM, yielding the image representation for the image generation task:
\[
\mathcal{V}_m = \text{MLP}(\text{Pool}_{a \times a}(\varphi(\mathcal{I}))) \in \mathbb{R}^{m \times d}.
\]


\subsection{Mutimodal Large Language Model}

To effectively manage both image understanding and image generation tasks, a carefully designed training objective is essential.
Given an image-text pair, for image understanding, we aim to predict the corresponding text based on the image. Conversely, for image generation, the text is used as a conditional prompt to generate the image.
Specifically, given a text, the embedding layer of the LLM maps each text token into a vector, forming the text embeddings $\mathcal{T}$. For image understanding, the image vectors $\mathcal{V}_n$ are prepended to the text embeddings to create \(\mathcal{X} = [\mathcal{V}_n ; \mathcal{T}]\), which is then fed into the LLM. The training objective for image understanding is to predict the probability distribution of the next text token based on the input image vectors and previously generated text tokens.

For image generation, the image vectors $\mathcal{V}_m$ are appended to the text embeddings $\mathcal{T}$, forming \(\mathcal{X} = [\mathcal{T} ; \mathcal{V}_m]\). In this task, the objective is to predict the next continuous image vector based on the input text and previously generated image vectors.

Thus, we can lossely define the unified training objective for two tasks as:
\begin{equation}
    \mathcal{L} = - \sum^{\mathcal{D}} \sum_{i=1}^{N} P_{\theta}(x_i | x_1,x_2,...,x_{n-1}),
\end{equation}
 where $x_i$ represents either a discrete text token or a continuous image vector, $\theta$ represents the parameters of the multimodal large language model, $N$ denotes the sequence length, and $\mathcal{D}$ is the training dataset comprising both image understanding and image generation instances.
The unified objective is optimized with two types of loss functions: (1) for image understanding, the CrossEntropy loss reduces the discrepancy between the predicted probability distribution of text tokens and the ground truth distribution; and (2) for image generation, the mean-squared-error (MSE) loss minimizes the difference between the predicted image vectors and the ground truth image vectors generated by the image encoder.

\vspace{-1mm}
\subsection{Image Decoding with Diffusion Models}
Given the predicted visual vectors \(\mathcal{V}_m\), we use a conditional diffusion model as a decoder to reconstruct the image from these vectors. This conditional diffusion model is pretrained with a CLIP image encoder, generating the original image conditioned on the final-layer embeddings from the CLIP model. This approach has shown effectiveness in previous studies~\cite{Emu2, ge2024seed, dreamllm}. During pretraining, the image encoder remains frozen, while the diffusion model is updated.

\subsection{Inference}
During inference for image generation, the multimodal large language model predicts all \(m\) image vectors in an autoregressive manner. These vectors are then mapped back to the image encoder's vector space through the image-generation head and subsequently input into the diffusion model to guide the denoising process. Following \cite{Emu2}, classifier-free guidance is provided by the multimodal model using an empty caption as the input. 
For image understanding, the process follows standard language modeling, using next-token prediction to generate textual outputs.

\section{Model Training and Data}

\subsection{Training Stage 1: Multimodal Pretraining}
In the pretraining stage, we simultaneously optimize the model for both image captioning and image generation. For image generation, we use 150 million high-quality image-caption pairs from the in-house Shutterstock dataset, applying the prompt ``Please generate an image based on the following caption: $<$caption$>$[IMG]$<$image$>$[/IMG]''. For image understanding, we use the same 150 million images, but with detailed captions generated by the Llama 3.2 model~\footnote{https://ai.meta.com/blog/llama-3-2-connect-2024-vision-edge-mobile-devices/}. The prompt used in captioning is ``[IMG]$<$image$>$[/IMG] Provide a detailed description of the given image. $<$caption$>$''.

\subsection{Training Stage 2: Fine-Grained Multimodal Pretraining}
In the fine-grained multimodal pretraining stage, we continue pretraining \modelname{} on 70 million image and detailed caption pairs for both generation and understanding. 
Similar to the detailed captions used in the first stage, these detailed captions are generated by the Llama 3.2 model. This stage aims to enhance the alignment between textual tokens and visual features in generated images while avoiding performance degradation in image understanding. The prompts for image generation and understanding remain same as the prompts used in the first stage. 

\subsection{Training Stage 3: Instruction Tuning for Image Understanding and Generation}
In the instruction tuning stage, we further refine the capabilities and generalizability of \modelname{} for both image understanding and generation. For image understanding, we carefully curated a large set of image-understanding tasks sourcing from two comprehensive datasets, including Cambrian-7M \cite{tong2024cambrian} and Vision Flan \cite{xu2024visionflanscalinghumanlabeledtasks}. We end up with 8 million high-quality images-text pairs. 
For image generation, we randomly sample 4 million image and short caption pairs, and 4 million image and detailed caption pairs from the Shutterstock dataset used in pretraining. By mixing instances with short and long captions, we preserve a broader distribution of input formats. The image generation and understanding datasets are merged, and training instances are sampled randomly at each step to promote a balanced and effective learning process.

\subsection{Implement Details}
In \modelname{}, we initialize the multimodal language model with LLaMA-3.1-Instruct 8B~\cite{llama3.1}, employ siglip-so400m-patch14-384~\cite{siglip} as the vision encoder for image understanding, and use a CLIP model trained with both MAE reconstruction loss~\cite{MAE} and contrastive loss~\cite{clip} as the vision encoder for image generation, denoted as gen-CLIP. Both the understanding and generation projection layers consist of two-layer MLPs, and the image-generation output head is also a two-layer MLP. For image decoding, we follow \citep{Emu2} to train SDXL~\cite{sdxl} as an image decoder for reconstructing images from gen-CLIP image embeddings. We leverage a 4x4 pooling kernel to pool the gen-CLIP embeddings into 64 continuous vectors.
In all three stages, we train the multimodal LLM, two MLPs for image understanding and generation while keeping both image encoders frozen. We set learning rate to 2e-5 and use constant with warm-up learning rate scheduler. The warm-up ratio is 0.03. The batch size in the first two stages is 2048 and in the third stage is 1024.

%% file: sec/3_experiment.tex

\section{Evaluation}

\begin{table*}[h!]
    \centering
    \begin{subtable}[t]{\textwidth}
        \centering
        \fontsize{8pt}{8.8pt}\selectfont
        \setlength\tabcolsep{8pt}
        \renewcommand{\arraystretch}{1.1}
        \resizebox{\textwidth}{!}{
        \begin{tabular*}{\linewidth}{l|cccccccccccc}
         Method & \rotatebox{90}{VQAv2} & \rotatebox{90}{GQA} & \rotatebox{90}{MMBench (EN)} & \rotatebox{90}{MMBench (CN)} & \rotatebox{90}{VizWiz} & \rotatebox{90}{POPE} & \rotatebox{90}{MM-Vet} & \rotatebox{90}{MME-P} & \rotatebox{90}{MME-C} & \rotatebox{90}{HallusionBench} & \rotatebox{90}{LLaVA-bench} & \rotatebox{90}{MMStar} \\
         \hline
        \rowcolor{gray!10}LLaVA 1.5 7B & 76.6 & 62.0 & 64.8 & 57.6 & 50.0 & 85.9 & 30.6 & 1510.7 & 294.0  & 44.8 & 64.2 & 30.3 \\
        \rowcolor{gray!10}LLaVA 1.5 13B & 78.3 & 63.3  & 68.7 & 62.5 & 56.7 & 85.9 & 35.3 & 1522.6 & 295.4 & 42.3 & 72.8 & 30.3 \\	
        EMU2 Chat 34B & - & - & 52.8 & - & - & \textbf{88.0} & 31.0 & - & - & 29.5 & 56.4 & 40.7  \\
       Chameleon 7B & - & - & 19.8 & - & - & 19.4 & 8.3 & 202.7 & - & 17.1 & 26.6 & 31.1\\
        Chameleon 34B & - & - & 32.7 & - & - & 59.8 & 9.7 & 604.5 & - & 18.6 & 33.3 & 31.8 \\
        Seed-X 17B & - & 47.9 & - & - & - & - & - & 1457.0 &  321.0 & - & - & - \\
        CM3Leon-7B & 47.6 &-&-&-&37.6 & - & - & - & -& -& -&- \\
        DreamLLM-7B & 72.9 & - & 58.2 & - & 49.3 & - & 36.6 & - & - & - & -  & - \\
        Show-o 1.3B & 73.9 & 47.5 & - & - & - & 73.8 & - & 1182.7 &  321.0 & - & - & - \\
        EMU3 8B & 75.1 & 60.3 & 58.5 & - & - & 85.2 & 37.2 & - & - & - & - & - \\
        Janus 1.3B & 77.3 & 59.1 & 69.4 & - & -& 87.0 & 34.3&1338.0 & - & -& -\\
        \rowcolor{green!10}\modelname{} 8B (Ours) & \textbf{82.1} & \textbf{64.8} & \textbf{73.9} & \textbf{65.2} & \textbf{72.1} & 86.3 & \textbf{50.0} & \textbf{1582.8} & \textbf{324.3} & \textbf{53.4} & \textbf{85.3} & \textbf{46.1}  \\
        \end{tabular*}
        }
        \caption{Results on general multimodal benchmarks.}
        \label{tab:ablation_method_1}
    \end{subtable}%
    \vspace{1em} 
    \begin{subtable}[t]{\textwidth}
        \fontsize{8pt}{8.8pt}\selectfont
        \centering
        \setlength\tabcolsep{9.9pt}
        \renewcommand{\arraystretch}{1.1}
        \resizebox{\textwidth}{!}{
        \begin{tabular*}{\linewidth}{l|ccc|cccc|ccccc}
        & \multicolumn{3}{c}{Vision centric} & \multicolumn{4}{c}{Knowledge based} & \multicolumn{4}{c}{OCR \& Chart}\\
          Method & \rotatebox{90}{Realworldqa} & \rotatebox{90}{CV-Bench*} & \rotatebox{90}{MMVP} & \rotatebox{90}{AI2D} & \rotatebox{90}{MathVista} & \rotatebox{90}{MMMU} & \rotatebox{90}{SciQA-IMG} & \rotatebox{90}{TextVQA} & \rotatebox{90}{OCRBench} & \rotatebox{90}{DocVQA} & \rotatebox{90}{InfoVQA} \\
        \hline
        \rowcolor{gray!10}LLaVA 1.5 7B  & 54.8 & 63.8  & 6.0 & 54.8 & 26.7 & 35.3 & 66.8 & 58.2 & 31.4 & 28.1 & 25.8 \\	
        \rowcolor{gray!10}LLaVA 1.5 13B & 55.3 & -  & - & 59.5 & 26.4 & 34.8 & 71.6 & 61.3 & 33.7 & 30.3 & 29.4 \\
        EMU2 Chat 34B & 57.3 & - & -& 49.7 & 30.7 & 35.7 & 68.2 & - & -& - & -  \\
        Chameleon 7B  & 39.0 & - & - & 46.0 & 22.3 & 22.4 & 46.8 & - & 0.0 &  - & -   \\
        Chameleon 34B & 39.2 & - & - & 53.7 & 23.6 & 38.8 & 58.8 & - & 0.0 &  - & -  \\
        Seed-X 17B & - & - & - & - & - & 35.6 & - & - & - & -  & -  \\
        DreamLLM-7B & - & - & - & - & - & - & - & 41.8 & - & -  & -  \\
        Show-o 1.3B & - & - & - & - & - & 27.4 & - & - & - & -  & -  \\
        EMU3 8B & 57.4 & - & - & 70.0 & - & 31.6 & - & 64.7 & -  & \textbf{76.3} & - \\
        Janus 1.3B & - & - & -& -& -& 30.5 & -& - & -& - &- \\
        \rowcolor{green!10}\modelname{} 8B (Ours) & \textbf{63.0} & \textbf{62.6} & \textbf{71.2} & \textbf{74.5} & \textbf{44.2} & \textbf{41.2} & \textbf{76.5} & \textbf{66.2} & \textbf{53.6} &  67.0 & \textbf{40.6}   \\
        \end{tabular*}%
        }
        \caption{Results on Vision centric, Knowledge based, and OCR \& Chart benchmarks.}
        \label{tab:und_results}
    \end{subtable}
    \caption{Results on general multimodal benchmarks, Vision centric, Knowledge based, and OCR \& Chart benchmarks. We highlight the best results in \textbf{bold}. LLaVA 1.5 7B and 13B are specialized image-understanding models.}
\end{table*}

\subsection{Image Understanding}
\textbf{Baselines}:
We evaluate our model against recent open-source unified models capable of both image understanding and generation, as well as against strong baselines specialized solely in image understanding. The unified models include EMU2 Chat \cite{Emu2}, Chameleon 7B and 34B \cite{chameleonteam2024chameleon}, Seed-X \cite{ge2024seed}, CM3Leon \cite{cm3leon}, DreamLLM \cite{dreamllm}, Show-o \cite{showo}, and EMU3 \cite{wang2024emu3nexttokenpredictionneed}. For comparison with specialized models, we use LLaVA 1.5 (7B and 13B)~\cite{liu2023improved}, which are dedicated to image understanding tasks.

\textbf{Benchmarks}: We evaluate the performance of different MLLM models on a comprehensive set of benchmarks with four different categories:
General multimodal benchmarks: VQAv2~\cite{goyal2017making}, GQA~\cite{hudson2019gqa}, MMBench (EN and CN)~\cite{liu2023mmbench}, VisWiz~\citep{gurari2018vizwiz}, POPE~\cite{li2023evaluating}, MM-Vet~\citep{yu2023mm}, MME Perception~\cite{fu2024mme}, MME Cognition~\cite{fu2024mme}, SeedBench~\citep{li2023seed}, HallusionBench~\cite{liu2023hallusionbench},  LLaVA in the Wild~\cite{liu2024visual} and MMStar~\cite{chen2024we}.
OCR \& Chart benchmark: TextVQA~\cite{singh2019towards}, OCRBench~\cite{liu2024hidden}, DocVQA~\cite{mathew2021docvqa} and InforVQA~\cite{mathew2022infographicvqa}.
Knowledge based benchmark: AI2D~\cite{kembhavi2016diagram}, MathVista~\cite{lu2023mathvista}, MMMU~\cite{yue2024mmmu} and ScienceQA~\cite{lu2022learn}.
Vision Centric benchmark: MMVP~\cite{tong2024eyes}, RealworldQA~\cite{grok15v} and CV-Bench~\cite{tong2024cambrian}.

\begin{table*}[!ht]
    \centering
    \resizebox{0.9\textwidth}{!}{
    \small
    \renewcommand{\arraystretch}{0.9}
    \begin{tabular*}{\linewidth}{@{\extracolsep{\fill}} l c c c c c c c}
        \toprule
        \textbf{Method} & \textbf{Single Obj.} & \textbf{Two Obj.} & \textbf{Counting} & \textbf{Colors} & \textbf{Position} & \textbf{Color Attri.} & \textbf{Overall} \\ 
        \midrule
        \multicolumn{8}{c}{\textit{Generation-Only Models}} \\
        \midrule
        LlamaGen~\citep{llamagen} & 0.71 & 0.34 & 0.21 & 0.58 & 0.07 & 0.04 & 0.32 \\ 
        LDM~\citep{LDM} & 0.92 & 0.29 & 0.23 & 0.70 & 0.02 & 0.05 & 0.37 \\ 
        SDv1.5~\citep{sdv2} & 0.97 & 0.38 & 0.35 & 0.76 & 0.04 & 0.06 & 0.43 \\ 
        PixArt-alpha~\citep{pixart} & 0.98 & 0.50 & 0.44 & 0.80 & 0.08 & 0.07 & 0.48 \\ 
        SDv2.1~\citep{sdv2} & 0.98 & 0.51 & 0.44 & 0.85 & 0.07 & 0.17 & 0.50 \\ 
        DALL-E 2~\citep{dalle2} & 0.94 & 0.66 & 0.49 & 0.77 & 0.10 & 0.19 & 0.52 \\ 
        SDXL~\citep{sdxl} & 0.98 & 0.74 & 0.39 & 0.85 & 0.15 & 0.23 & 0.55 \\ 
        SD3 (d=24)~\citep{SD3} & 0.98 & 0.74 & 0.63 & 0.67 & 0.34 & 0.36 & 0.62 \\ 
        \midrule
        \multicolumn{8}{c}{\textit{Understanding \& Generation Models}} \\
        \midrule
        CoDI~\citep{codi} & 0.89 & 0.16 & 0.16 & 0.65 & 0.02 & 0.01 & 0.31 \\ 
        LWM~\citep{discreteToken} & 0.93 & 0.41 & 0.46 & 0.79 & 0.09 & 0.15 & 0.47 \\ 
        SEED-X~\citep{ge2024seed} & 0.97 & 0.58 & 0.26 & 0.80 & 0.19 & 0.14 & 0.49 \\ 
        Chameleon~\citep{chameleonteam2024chameleon} & - & - & - & - & - & - & 0.39 \\ 
        Show-o~\citep{showo} & 0.95 & 0.52 & 0.49 & 0.82 & 0.11 & 0.28 & 0.53 \\ 
        Emu3~\citep{wang2024emu3nexttokenpredictionneed} & - & - & - & - & - & - & 0.64 \\
        Transfusion~\citep{zhou2024transfusionpredicttokendiffuse} & - & - & - & - & - & - & 0.63 \\
        Janus~\cite{wu2024janus} & 0.97 & 0.68 & 0.30& \textbf{0.84}& 0.46&\textbf{0.42}& 0.61 \\
        \rowcolor{green!10} \modelname{} (Ours) & \textbf{1.00}& \textbf{0.76}& \textbf{0.48} & 0.83 & \textbf{0.52}& 0.28 & \textbf{0.65} \\
        \bottomrule
    \end{tabular*}}
    \caption{Performance on the text-to-image GenEval Benchmark.}
    \label{tab:geneval}
\end{table*}

\subsection{Image Generation}
\textbf{Baselines}: 
We compare our model with state-of-the-art unified models capable of both image understanding and generation, including CoDI~\cite{codi}, LWM, SEED-X~\cite{ge2024seed}, EMU~\cite{emu}, Chameleon \cite{chameleonteam2024chameleon}, Transfusion~\cite{zhou2024transfusionpredicttokendiffuse}, Show-o~\cite{showo}, EMU3~\cite{wang2024emu3nexttokenpredictionneed}, and Janus~\cite{wu2024janus}. 
Additionally, we benchmark against models specialized in image generation, such as SDv1.5~\cite{sdv2}, DALL-E 2~\cite{dalle2}, PixArt-alpha~\cite{pixart}, Llama Gen~\cite{llamagen}, LDM~\cite{LDM}, SDv2.1~\cite{sdv2}, SDXL~\cite{sdxl}, and SDv3~\cite{SD3}. 

\textbf{Benchmarks}: We assess the image generation performance of \modelname{} on GenEval~\cite{geneval}, a widely adopted benchmark for image generation. We leverage the official implementation of the evaluation metric for GenEval~\footnote{https://github.com/djghosh13/geneval}.

%% file: sec/4_results.tex

\section{Main Results}

\subsection{Image-Understanding}


\textbf{Comprehensive Evaluation Benchmarks}: 

As shown in Table \ref{tab:und_results} (a), \modelname{} achieves the state-of-the-art performance on most of the comprehensive evaluation benchmarks compared to open-source unified models. It even outperforms unified models 2 or 4 times larger such as Seed-X 17B and Chameleon 34B, highlighting its strong capability as a general-purpose visual-chat assistant. In addition, our model surpass all recent models by a large margin, such as 26.3\% on MMBench comparing to EMU3, 8.6\% on MME-P comparing to Seed-X, and 34.4\% on MM-Vet comparing to EMU3.

\textbf{Vision centric, Knowledge based, and OCR \& Chart Benchmarks}: We show the results of \modelname{} on domain specific benchmarks in Table \ref{tab:und_results} (b). As one can observe, our model achieves the best performance on most of the tasks comparing to open-source unified models, and even achieves comparable performance compared to strong models specialized in understanding.

\subsection{Conditional Image Generation}
\begin{figure*}[h!]
  \centering
   \includegraphics[width=0.8\linewidth]{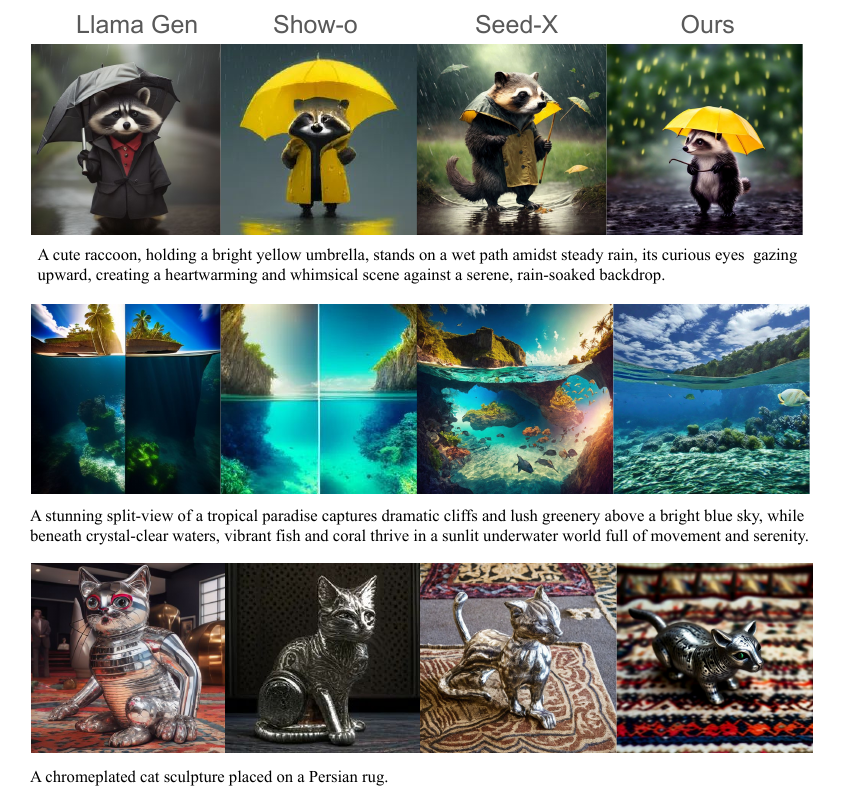}
   \caption{Qualitative comparison of image generation results.}
   \label{fig:qualitative}
\end{figure*}

We present the performance of \modelname{} on the GenEval benchmark in Table \ref{tab:geneval}. Notably, \modelname{} achieves the competitive performance among the unified understanding and generation models on GenEval, highlighting its strong instruction-following capabilities in image generation. We also present a qualitative comparison between \modelname{} and unified models with image generation capability in Figure~\ref{fig:qualitative}. As one can observe, \modelname{} can follow complex user prompts to generate high-quality images.

\section{Discussion}
\subsection{Synergistic Relationship between Image Understanding and Image Generation}\label{sec:gen_and_und}
\begin{table*}[h!]
    \centering
    \resizebox{0.9\textwidth}{!}{
    \renewcommand{\arraystretch}{1.0}
    \begin{tabular*}{\linewidth}{@{\extracolsep{\fill}} l c c c c c c c}
    \toprule
    {\textbf{Methods}} & \textbf{Pope} & \textbf{AI2D} & \textbf{MME} & \textbf{MathVista} & \textbf{MM-Vet} & \textbf{SeedBench} & \textbf{MSCOCO-30K FID$\downarrow$} \\
    \midrule
    \modelname{} & \textbf{87.7} & \textbf{64.1} & \textbf{1507.0} & \textbf{31.3} & \textbf{37.5} & \textbf{74.4} & \textbf{38.2} \\ 
    \modelname{} w/o Und. & - & - & - & -& -&- & 78.4  \\
    \modelname{} w/o Gen.  & 86.2 & 60.7 & 1460.2  & 28.0 & 32.9 & 68.4 & - \\
    \bottomrule
    \end{tabular*}}
    \caption{Performance of \modelname{} without training on the image generation task and without training on the image understanding task.}
    \label{tab:und_vs_gen}
\end{table*}

In this section, we examine within the unified model architecture whether image understanding and image generation can mutually benefit. Due to limited computational resources, we sample a subset of 15 million image and detailed caption pairs used in training stage 1 for pretraining.
For instruction tuning, we utilize the LLaVA-Instruct-150K dataset~\cite{liu2023improved}. We train three model variants under different data settings: (1) PISCES, which is pretrained on a combination of image understanding and generation data, subsequently instruction-tuned on LLaVA-Instruct-150K; (2) PISCES w/o Und, pretrained exclusively on the Shutterstock dataset; and (3) PISCES w/o Gen, pretrained on the PixelProse dataset and finetuned on LLaVA-Instruct-150K. We train the multimodal LLM, two MLPs for image understanding and generation while keeping both image encoders frozen. We set learning rate to 2e-5 and use cosine decay with warm-up learning rate scheduler. The warm-up ratio is 0.03. The batch size is 1024.

We evaluate the performance of these three variants on a comprehensive set of image understanding benchmarks and measure their FID scores on MSCOCO-30K~\cite{mscoco}, as reported in Table~\ref{tab:und_vs_gen}. It is evident that, under the constrained training data setting, image understanding substantially enhances image generation, and conversely, image generation contributes to improved image understanding. This result substantiates the synergistic relationship between image understanding and generation.

\subsection{Benefits of Decoupled Vision Encoders}\label{abl:decouple_encoder}

We perform ablation study to verify the effectiveness of using decoupled visual encoders in two settings: (1) we use gen-CLIP for both image understanding (without pooling totally 1024 visual tokens) and image generation (with pooling totally 64 visual tokens) and follow the exact training setting as described in Section~\ref{sec:gen_and_und} to train \modelname{}. The resulted \modelname{}-gen-CLIP model achieves on par performance as \modelname{} with decouple encoder on image generation (40.5 of \modelname{}-gen-CLIP vs. 38.2 of \modelname{}) but inferior image understanding performance shown in Table~\ref{tab:decouple}; (2) we also tried to use SigLIP for both image understanding and image generation.
\begin{wraptable}{r}{0.52\textwidth}
    \centering
    \small
    \renewcommand{\arraystretch}{1.0}
    \setlength{\tabcolsep}{2.5pt}
    \caption{Performance of \texttt{\modelname{}} with decoupled image understanding encoder (SigLip) and image generation encoder (gen-CLIP), and \texttt{\modelname{}} with the same image understanding and generation encoder (gen-CLIP).}
    \vspace{0.5em}
    \resizebox{0.48\textwidth}{!}{
    \begin{tabular}{l c c c c c c}
        \toprule
        \textbf{Methods} & \textbf{Pope} & \textbf{AI2D} & \textbf{MME} & \textbf{MathVista} & \textbf{MM-Vet} & \textbf{SeedB.} \\
        \midrule
        Decoupled & \textbf{87.7} & \textbf{64.1} & \textbf{1507.0} & \textbf{31.3} & \textbf{37.5} & \textbf{74.4} \\
        gen-CLIP  & 80.8 & 48.7 & 1185.5 & 22.2 & 27.3 & 62.8 \\
        \bottomrule
    \end{tabular}}
    \label{tab:decouple}
\end{wraptable}
We train the SDXL model to decode SigLIP features using 30 million high-quality images from Shutterstock. However, the SigLIP+SDXL architecture does not achieve the same level of reconstruction performance as the Gen-CLIP+SDXL architecture. We attribute this discrepancy to the absence of a masked autoencoding (MAE) loss during SigLIP’s pretraining, which limits the model’s ability to capture fine-grained pixel-level details. This limitation underscores the benefit of decoupling image encoders for understanding and generation, as the most effective publicly available encoders for these tasks are often not the same model.

\subsection{Effect of Number of Visual Tokens for Image Generation}
\label{abl:num_token}

\begin{wrapfigure}{r}{0.45\linewidth}
  \centering
  \includegraphics[width=\linewidth]{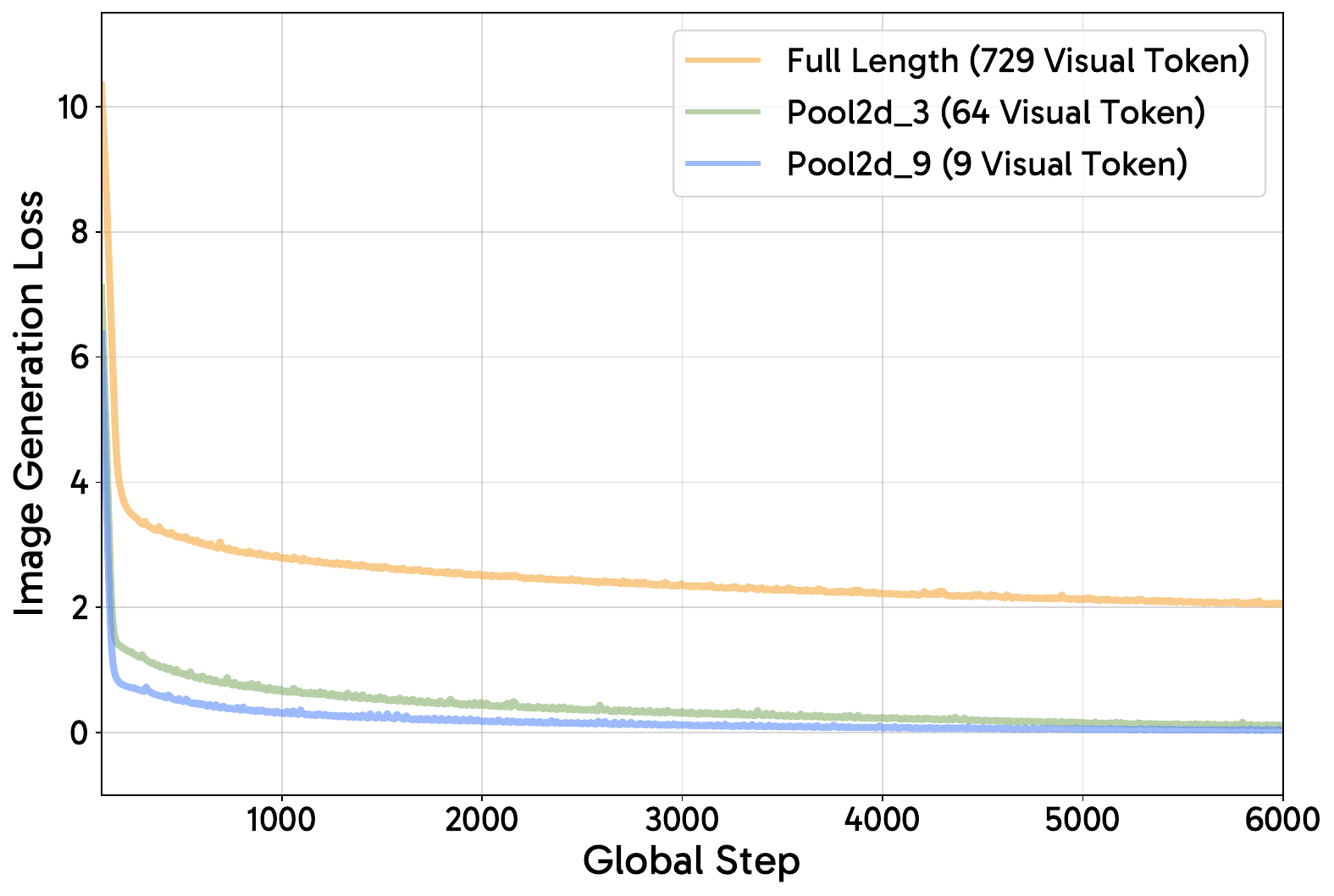}
  \caption{Image generation loss varies with the number of visual tokens, with a higher number of visual tokens resulting in increased training loss, highlighting the challenges faced during LLM training.}
  \label{fig:generation_loss}
\end{wrapfigure}

In the image generation task, the LLM is trained to predict continuous image vectors in the autoregressive manner. However, we observe that longer sequence lengths present a significant challenge for the LLM. To quickly explore the number of image tokens required for image generation, we conduct experiments using SigLIP ViT-L/16@384. Specifically, we compare the full sequence length of 729 image tokens with pooling strategies: a stride of 3, yielding 81 visual tokens, and a stride of 9, yielding 9 visual tokens. For the image generation task, we train three models with different number of input image tokens for 6,000 steps. As shown in Figure \ref{fig:generation_loss}, using the full length (729 visual tokens) yields the highest training loss. Reducing the number of visual tokens lowers the loss, although using only 9 visual tokens achieves the lowest generation loss but leads to poor performance due to excessive information loss during pooling. Therefore, we select the intermediate approach, pooling with a stride of 3.

\subsection{Effect of Detailed Captions on Image-Generation} \label{abl:second_stage}
In this section, we illustrate the significance of our second-stage pretraining on detailed captions.
We take the \modelname{} from the first-training stage, which is trained on 150 million of image and short caption pairs for generation, as the starting point and perform further pretraining in two distinct settings. In the first setting, we conduct second-stage pretraining using 70 million detailed image captions to enhance both image generation and understanding.

\begin{wraptable}{h}{0.4\textwidth}
    \centering
    \small
    \renewcommand{\arraystretch}{1.0}
    \caption{Performance comparison of \modelname{} trained on short captions versus long captions in the second-stage training.}
    \vspace{0.5em}
    \begin{tabular*}{\linewidth}{@{\extracolsep{\fill}} l c}
        \toprule
        \textbf{Training Stage} & \textbf{MSCOCO-30K $\downarrow$} \\
        \midrule
        1st-Stage & 25.6 \\
        2nd-Stage w/ Short Captions & 27.6 \\
        2nd-Stage w/ Long Captions & \textbf{21.9} \\
        \bottomrule
    \end{tabular*}
    \label{tab:long_vs_short}
\end{wraptable}

In the second setting, the model is pretrained on 70 million short image captions for image generation and 70 million detailed captions for image understanding. We present the FID scores of these models on MSCOCO-30K in Table~\ref{tab:long_vs_short}. As observed, incorporating more short captions does not yield further improvement in the model's FID, whereas using long captions consistently enhances image generation performance.

%% file: sec/5_related_work.tex
\section{Related Work}

Recently works \cite{liu2023llava,li2023blip} demonstrate that by bootstrappning pretraining from pretrained language models and vision encoders, one can easily train a strong image-understanding model with moderate computation cost. \citep{xu-etal-2023-multiinstruct} proposes the first human-label multi-modal instruction tuning dataset and later works \citep{ye2023mplug, yin2023lamm, mimic,lyu2023macaw, zhu2023minigpt, Dai2023InstructBLIPTG, liu2023improved, chen2023sharegpt4vimprovinglargemultimodal, prompt_GPT4V, xu2024visionflanscalinghumanlabeledtasks} augment visual instruction tuning with human-labeled and GPT4-labeled instructions. More recently, \cite{bai2023qwenvl, liu2024llavanext, llava_interleave, mixLoRA, llava_UHD} propose various training techiques and model architectures to improve image-understadning models and allow more flexiable input image and text formats.
Two primary approaches are commonly used in unified multimodal models. The first approach utilizes VQGAN~\citep{vqgan} to discretize an image into a sequence of tokens, subsequently incorporating VQGAN’s codebook into the language model's vocabulary~\citep{cm3,cm3leon,recm3,chameleonteam2024chameleon,discreteToken}. This allows the language model to train under a unified autoregressive objective, predicting either image tokens or text tokens. 
The second approach employs the CLIP image encoder, which encodes images as sequences of continuous embeddings~\citep{gill,codi2,vlGPT,emu,Emu2,minigemini,nextGPT,mmInterleave}. These embeddings are combined with text embeddings in their sequential order, typically allowing for shorter sequence lengths and yielding improved performance over the token-based approach.
However, both the first and the second approaches utlize the same vision encoder, projection layers, and number of visual tokens for image generation and understanding, ignoring the intrinsic discrepency between two tasks and yeilding suboptimial performance. 

%% file: sec/6_conclusion.tex
\section{Conclusion}
In this paper, we presented \modelname{}, a unified multimodal foundation model that effectively bridges the gap between image understanding and generation. By introducing an asymmetrical visual encoding architecture and task-specific training techniques, \modelname{} demonstrates that a single model can achieve strong performance across diverse modalities without sacrificing the depth of specialization. Our extensive evaluations across over 20 benchmarks validate the model’s capability to perform at the forefront of both image understanding and generation tasks. We believe that the insights gained from \modelname{} will inspire future innovations, driving more robust and versatile approaches to multimodal modeling.